\documentclass[preprint, conference]{IEEEtran}

\IEEEoverridecommandlockouts
\usepackage{cite}
\usepackage{amsmath,amssymb,amsfonts}
\usepackage{algorithmic}
\usepackage{graphicx}
\usepackage{textcomp}
\usepackage{{booktabs}}
\usepackage{xcolor}

\def\BibTeX{{\rm B\kern-.05em{\sc i\kern-.025em b}\kern-.08em
    T\kern-.1667em\lower.7ex\hbox{E}\kern-.125emX}}
\begin{document}

\title{Ask-EDA: A Design Assistant Empowered by LLM, Hybrid RAG and Abbreviation De-hallucination}

\author{
\IEEEauthorblockN{Luyao Shi}
\IEEEauthorblockA{\textit{IBM Research} \\
San Jose, CA \\
luyao.shi@ibm.com}
\and
\IEEEauthorblockN{Michael Kazda}
\IEEEauthorblockA{\textit{IBM Infrastructure} \\
Poughkeepsie, NY \\
kazda@us.ibm.com}
\and
\IEEEauthorblockN{Bradley Sears}
\IEEEauthorblockA{\textit{IBM Infrastructure} \\
Austin, TX \\
bradley@us.ibm.com}
\and
\IEEEauthorblockN{Nick Shropshire}
\IEEEauthorblockA{\textit{IBM Infrastructure} \\
Austin, TX \\
nick.shropshire@ibm.com}
\and
\IEEEauthorblockN{Ruchir Puri}
\IEEEauthorblockA{\textit{IBM Research} \\
Yorktown Heights, NY \\
ruchir@us.ibm.com}
}

\maketitle

\begin{abstract}
Electronic design engineers are challenged to find relevant information efficiently for a myriad of tasks within design construction, verification and technology development.  Large language models (LLM) have the potential to help improve productivity by serving as conversational agents that effectively function as subject-matter experts. In this paper we demonstrate Ask-EDA, a chat agent designed to serve as a 24$\times$7 expert available to provide guidance to design engineers. Ask-EDA leverages LLM, hybrid retrieval augmented generation (RAG) and abbreviation de-hallucination (ADH) techniques to deliver more relevant and accurate responses. We curated three evaluation datasets, namely q2a-100, cmds-100 and abbr-100. Each dataset is tailored to assess a distinct aspect: general design question answering, design command handling and abbreviation resolution. We demonstrated that hybrid RAG offers over a 40\% improvement in Recall on the q2a-100 dataset and over a 60\% improvement on the cmds-100 dataset compared to not using RAG, while ADH yields over a 70\% enhancement in Recall on the abbr-100 dataset.  The evaluation results show that Ask-EDA can effectively respond to design-related inquiries.

\end{abstract}

\begin{IEEEkeywords}
Large language models, LLM, design assistant, chatbot, EDA, retrieval augmented generation, RAG, hybrid search, de-hallucination
\end{IEEEkeywords}

\section{Introduction}
Modern design engineers face formidable challenges. They are confronted with a variety of tasks for design construction and verification. The process of locating the correct document or a subject-matter expert to consult is frequently a problem in large organizations. There can be redundant document versions and sometimes they are not located in centralized locations. This problem is especially important for newly hired employees or with new tools being brought online. There can be a new set of jargon and acronyms that must be understood. Having a 24$\times$7 consultant available to advise engineers would significantly enhance productivity.

Large language models (LLMs) \cite{achiam2023gpt} are exceptionally good at providing natural language responses. Despite the impressive performance of LLMs across various tasks, their responses are constrained by the limitations of their training data. The training data for LLMs often becomes outdated, and we aim to avoid incorporating confidential design information into it. When queried about topics outside its training data, an LLM might confidently produce incorrect yet seemingly plausible responses, a phenomenon known as hallucination. Retrieval-Augmented Generation (RAG) \cite{lewis2020retrieval} addresses these concerns by combining information retrieval with thoughtfully crafted system prompts. This method anchors LLMs to precise, current, and relevant information retrieved from an external knowledge repository.

Sentence transformer \cite{reimers2019sentence} is one of the state-of-the-art information retrieval models, and has quickly becomes a popular choice for RAG \cite{liu2023chipnemo}. A sentence transformer encodes text into dense vector representations, capturing semantic information and enabling more accurate retrieval based on meaning rather than just keyword matching. While sentence transformers-based dense retrieval methods excel at retrieving relevant semantic context, there are instances where design engineers seek results containing specific technical terms, which dense retrieval methods cannot consistently guarantee to retrieve. Alternatively, sparse retrieval methods \cite{robertson2009probabilistic, formal2021splade, gao-etal-2021-coil} allow for exact matching of query terms with document terms, enabling precise retrieval based on exact term matches. We developed a hybrid search engine that leverages the strengths of both dense and sparse retrieval algorithms to improve the accuracy and relevance of search results.

Another challenge we encounter with LLMs is their tendency to generate hallucinated explanations for abbreviations when they lack knowledge of the correct full names. This phenomenon is particularly prevalent in the design space, where abbreviations are commonly utilized. To tackle this, we derived an Abbreviation De-Hallucination (ADH) component, which utilizes a pre-built abbreviation dictionary to furnish relevant abbreviation knowledge to LLMs.

In this work, we developed Ask-EDA, a chat agent designed to support Electronic Design Automation (EDA) and enhance productivity for design engineers. Ask-EDA leverages LLM, hybrid RAG, and abbreviation de-hallucination techniques to deliver more relevant and accurate responses. We curated three evaluation datasets to demonstrate the effectiveness of Ask-EDA, namely q2a-100, cmds-100 and abbr-100, which focus on the aspects of general design question answering, design command answering and abbreviation resolution, respectively. While prior arts such as ChipNeMo \cite{liu2023chipnemo} has demonstrated the capability of  a design assistant chatbot, our incorporation of hybrid RAG and ADH components demonstrates a notable enhancement in response quality, irrespective of the LLM models employed. Additionally, we extend our evaluation to encompass diverse sets that address various challenges in question answering within the design domain. We also use Slack API \cite{slackboltapi} to build a natural language interface that allows users to make conversations with our chat agent.

\section{Methodology}

\subsection{Document Sources}
There are several sources of documents which our design team use as a reference. Subject matter experts (SMEs) from the design team author guidance for a given chip and methodology. Our foundry suppliers provide technology physical design kit manuals specifying design rules. The tool and methodology SMEs provide comprehensive tool documentation encompassing commands, parameters, and methodology steps. DevOp SMEs provide documents on job submission and continuous integration procedures.  Slack provides a wealth of conversations on procedures and contact points.   Designers and developers alike use engineering workflow management for iteration and release planning, change management, and defect tracking. Internally, we have our own stack-overflow-type interface and database for asking questions, up-ranking the correct answer, and information retrieval.  Presently, the sources consume about 400Mb on disk and represent approximately 10,200 command manual pages, 5,000 parameters, 30 slack channels, and 18,000 commonly asked questions/answers.

\subsection{Hybrid RAG}
We use Retrieval Augmented Generation (RAG) to provide relevant context for LLM generation. We developed a hybrid search engine to combine sentence transformers and BM25 to improve the accuracy and relevance of search results. Below are details regarding both ingestion and retrieval phases.

\subsubsection{Ingestion}
\begin{figure}[htbp]
    \centering
    \includegraphics[width=0.47\textwidth]{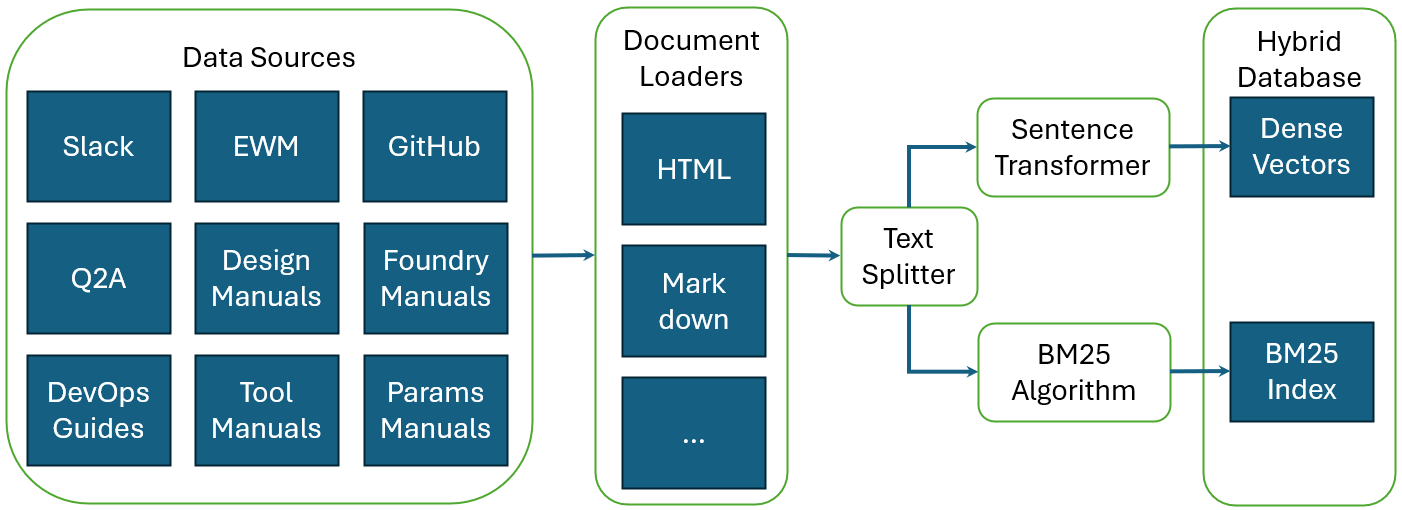}
    \caption{Document ingestion into hybrid database.}
    \label{fig:ingestion}
\end{figure}

Each of the source documents can be in a different format, therefore, we utilize langchain \cite{langchain} document loaders to read in the documents in Fig.\ref{fig:ingestion}. This supports comma and tab separated value files, json, pdf, docx, pptx, markdown, and plain text formats.  The documents are chunked into evenly sized chunks presently.  Each chunk is fed to a sentence transformer to create a dense embedding vector. We use ChromaDB \cite{chroma} as our dense vector database. The same chunk is also fed to BM25 \cite{robertson2009probabilistic} to calculate BM25 index. Together they form a hybrid database that will be used for retrieval later.

\subsubsection{Retrieval}
When the user makes a query,  we encode the query using the same sentence transformer at ingestion phase and match its embedding to the nearest embeddings in the dense vector database with cosine similarity. This provides us the top $n_{dense}$ most semantically-relevant text chunks. We also find the most-relevant $n_{sparse}$ text chunks based on the pre-computed BM25 index. The results of the dense and sparse methods are then combined with reciprocal rank fusion (RRF) \cite{cormack2009reciprocal}. The RRF score is calculated as:
  
\begin{equation}
RRFscore(d \in D) = \sum_{r \in R}\frac{1}{k + r(d))}
\label{eq:rrf}
\end{equation}

where $D$ is the union of the top $n_{dense}$ and top $n_{sparse}$ candidate text chunks and $d$ is one text candidate. $r(d)$ is the ranking index from one ranking method (either dense or sparse method sorted based on its relevance scores), and $R$ is the set of rankings from different ranking methods. $k$ is a constant that helps to balance between high and low ranking and is set to 60 in our study. The set of text chunks will be re-ranked based on their RRF scores, and the top $n_{hybrid}$ text chunks with highest RRF scores will be selected as context in the LLM prompt. These text chunks are sorted in ascending order so that the most relevant context is closer to the user query in the LLM prompt.

\begin{figure}[htbp]
    \centering
    \includegraphics[width=0.47\textwidth]{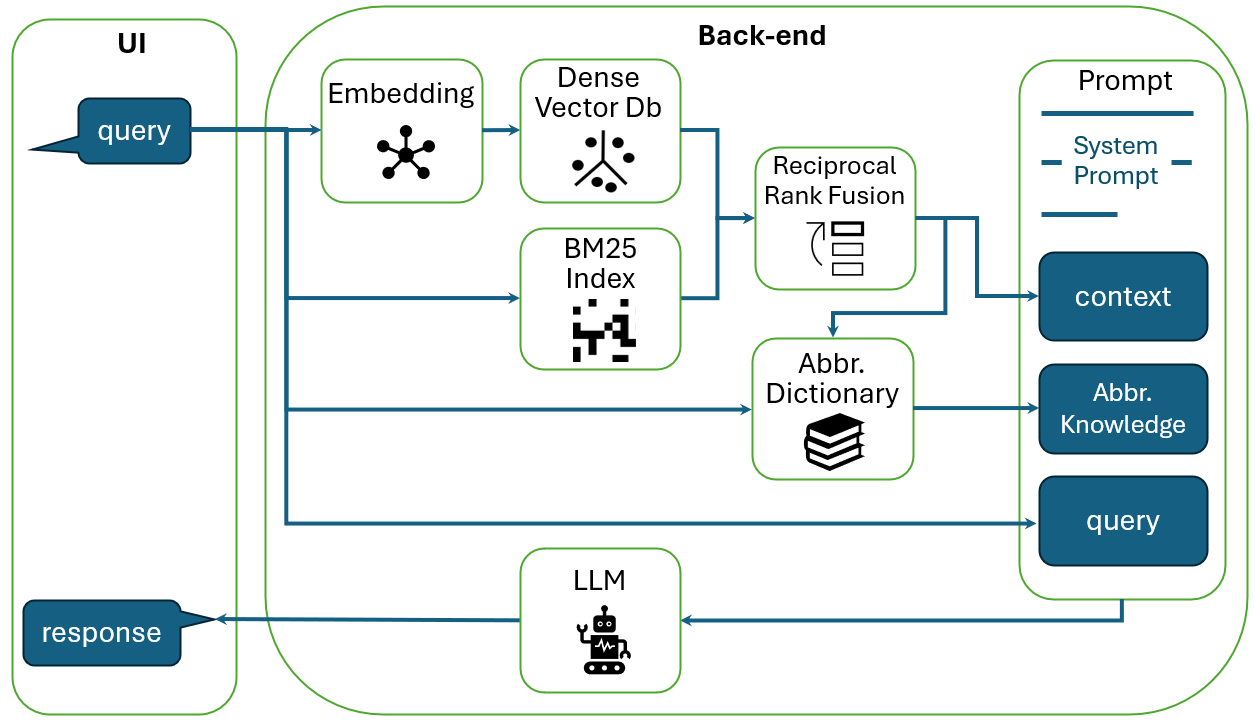}
    \caption{Pipeline to produce a response based on user query.}
    \label{fig:retrieval}
\end{figure}

\subsection{Abbreviation De-hallucination}
To help alleviate the production of hallucinated responses by LLM regarding abbreviations, we created an Abbreviation (Abbr.) Dictionary, which contains 249 common design-related abbreviation terms. These terms are predominantly specific to IBM chip design, with approximately 25\% being commonly recognized in the electronic design automation industry. All abbreviation items have their full names and 148 out of 249 terms have additional detailed descriptions. Five design experts contributed to building this dictionary. When the user makes a query, the relevant context is first obtained through hybrid search and RRF. We then search through both the query and the context to find relevant abbreviations in the dictionary based on exact term matching. If relevant abbreviations are found, their associated full names and descriptions (if available) are pulled out from the dictionary and added to the prompt to provide abbreviation knowledge for LLM. The abbreviation knowledge is added in between context and query in our implementation. For each abbreviation term, its knowledge is added in the format of ``$abbr$ is usually short for $name$, which is $desc$.'' when its description is available, and ``$abbr$ is usually short for $name$.'' when its description is absent, where $abbr$ is the abbreviation term, $name$ is its full name and $desc$ is its description.

\subsection{LLM Generation}
We finally add a system prompt at the beginning of our prompt, to direct the behavior of LLM and ensure that the generated outputs align with our intended goals. The system prompt that we used will be provided in subsequent sections of the paper. This final prompt is input to the LLM and a response is generated. The LLM response along with the relevant documents are returned to the user as a complete response. Our pipeline is presented as a diagram in Fig. \ref{fig:retrieval}.

\subsection{Chat Interface}
Slack is commonly used by engineers for communication and is a natural interface for holding conversations. We have created an agent through the Slack API. The user sends a message to the agent and it will create a response in a thread. Chat history continues the conversation and we include historical context from the most recently prior questions for subsequent queries. We also provide a interface that enables users to review the sources provided by RAG and give feedback on the quality of Ask-EDA’s response. Note that chat history and the feedback data are not used in the evaluation study in this paper. An example screenshot of our interface is shown in Fig. \ref{fig:interface}.

\section{Evaluation}

\subsection{Datasets}
To evaluate the performance of our Ask-EDA, we have curated three datasets for evaluation, namely q2a-100, cmds-100 and abbr-100. The content of each is domain-specific to IBM chip design and methodology.

\subsubsection{q2a-100 Dataset}
We have a stack-overflow type system where engineers are able to pose a question and other SMEs can comment and answer. There is a voting system which the best answer can be marked. Using these expert answers to common questions represent a collective knowledge with ground truth labeled data. We have extracted those questions and answers as a data source to be ingested. Based on this database, we have created a subset containing 100 questions and answers as an evaluation dataset.

\subsubsection{cmds-100 Dataset}
The tools team creates manual pages to document the commands of our physical construction and verification system. These are a toolkit where designers can augment the methodology flow for special needs. Each of these commands represent a data source from which developers frequently consult for the command and options. This content is HTML generated from the tool source code. We similarly have extracted this as a design source. We further collected 100 commands as a test dataset by turning the one-line synopsis of a command into a question and providing the command name as the ground truth answer.

\subsubsection{abbr-100 Dataset}
We derived the abbr-100 dataset as a subset from the abbreviation dictionary described previously. We randomly sample 100 abbreviation terms from the abbreviation dictionary, and created questions and answers in the format of ``What does $abbr$ stand for?'', ``$name$'', respectively. Note that in this evaluation set, only the abbreviations and their full names are included in the question-answer pairs, without considering the descriptions. In Table. \ref{tab:example} we show one example from each evaluation dataset described above.

\begin{table}[htbp]
\caption{Examples of question-answer pairs in the three evaluation datasets.}
\label{tab:example}
\centering
\begin{tabular}{ll}
\hline
Dataset  & Example  \\ \hline
q2a-100  & \begin{tabular}[c]{@{}l@{}}\textbf{Question}: How do I define a placement blockage region? \\ \textbf{Answer}: You can define placement blockage using these \\ two params. The first defines blockage which...\end{tabular} \\ \hline
cmds-100 & \begin{tabular}[c]{@{}l@{}}\textbf{Question}: What is the Tcl command that can get the pin \\ capacitance? \\ \textbf{Answer}: ess::get\_pin\_capacitance \end{tabular} \\ \hline
abbr-100 & \begin{tabular}[c]{@{}l@{}}\textbf{Question}: What does RAT stand for? \\ \textbf{Answer}: Required Arrival Time \end{tabular} \\ \hline
\end{tabular}%
\end{table}

\subsection{Implementation Details}
We evaluated two LLMs in our study. Granite-13b-chat-v2.1 \cite{ibm-2024-granite, mishra2024granite} is a decoder-only and English-only foundation model developed and trained by IBM. There is no overlap between the LLM training data and our ingested data. We also evaluated Llama2-13b-chat, which is considered to be one of the state-of-the-art open-source chat models at the 13b parameter scale. For both models, the context length is 8192 and the max new tokens is 4096. For Granite-13b-chat-v2.1 and Llama2-13b-chat, we use their corresponding prompt formats but use the same system prompt: ``You are a helpful AI language model. Your primary function is to assist users in answering questions, generating text, and engaging in conversation. Given the following extracted parts of a long document and a question, create a final answer. If asking for a command, please return the first one only.'' We use all-MiniLM-L6-v2 \cite{wang2020minilm} as text embedder. The chunk size at ingestion is 2048 with chunk overlap 256. The numbers of retrieval candidate $n_{dense}$, $n_{sparse}$ and $n_{hybrid}$ are all set to 3.

\begin{figure}[htbp]
    \centering
    \includegraphics[width=0.47\textwidth]{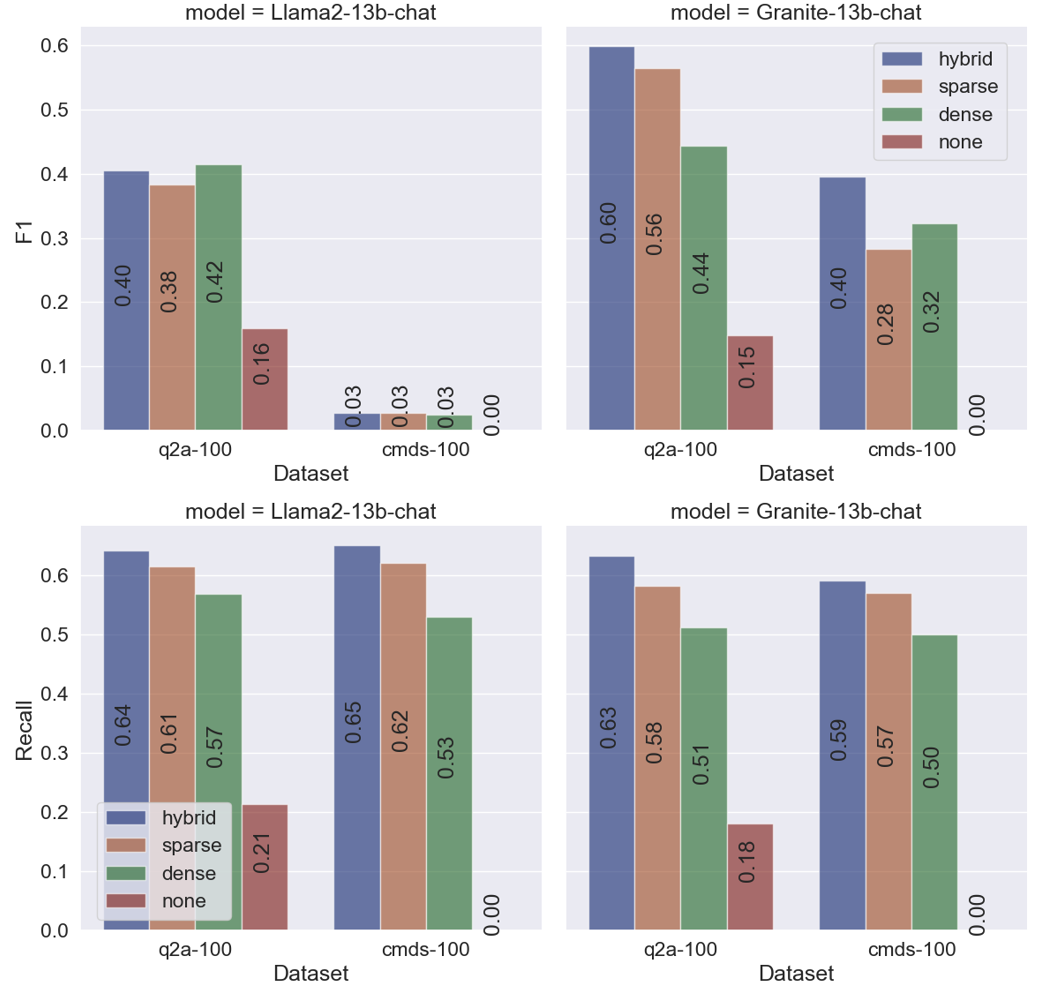}
    \caption{Results on q2a-100 and cmds-100 datasets.}
    \label{fig:charts1}
\end{figure}

\subsection{Results}
Fig. \ref{fig:charts1} shows the results of Granite-13b-chat-v2.1 and Llama2-13b-chat models on q2a-100 and cmds-100 datasets. For each LLM model, we compared different RAG techniques, including: hybrid retrieval (hybrid), BM25-only sparse retrieval (sparse), sentence transformer-only dense retrieval (dense) and no RAG (none). The abbreviation de-hallucination (ADH) component is not included at this point. We use ROUGE-Lsum \cite{lin-2004-rouge} F1 and Recall as our evaluation metrics. From Fig. \ref{fig:charts1} we can see that on q2a-100, Granite-13b-chat-v2.1 generates significantly better responses compared to Llama2-13b-chat in terms of F1 score, while providing comparable responses in terms of Recall. For both models, employing RAG yields significantly superior results compared to not using RAG. With Granite-13b-chat-v2.1, hybrid retrieval achieves the highest performance, followed by sparse and dense retrieval. However, with Llama2-13b-chat, this pattern is observed only in terms of Recall, not F1. Given its comparatively lower F1 performance, it is plausible that Llama2-13b-chat struggles to effectively extract the final answer from the context, despite the potential improvement offered by hybrid search.

On cmds-100 dataset, Llama2-13b-chat achieves slightly better Recall than Granite-13b-chat-v2.1, but Granite-13b-chat-v2.1 delivers vastly superior responses than Llama2-13b-chat in terms of F1 score. It's noticeable that the no RAG models achieve a Recall score of 0 since LLMs lack knowledge about these design commands. Here, hybrid search RAG once again outperforms both sparse-only and dense-only RAG.

\begin{figure}[htbp]
    \centering
    \includegraphics[width=0.47\textwidth]{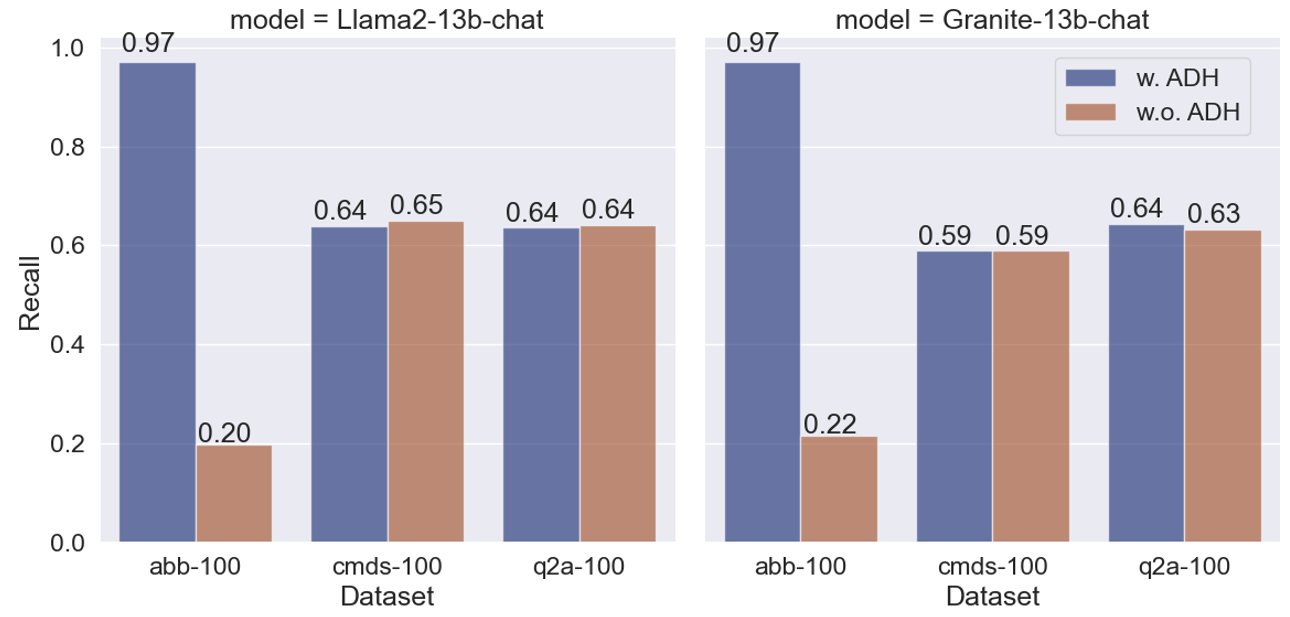}
    \caption{Evaluation results on abbreviation de-hallucination.}
    \label{fig:charts2}
\end{figure}

Fig. \ref{fig:charts2} shows the comparison results after adding the abbreviation de-hallucination (ADH) component. Only hybrid RAG is utilized here. We exclusively report the Recall performance here because the answers in abbr-100 consist of brief and clearly stated full abbreviation names and LLMs sometimes produce lengthy responses with descriptive explanations, which can lead to penalized F1 scores. It can be seen that the addition of ADH component significantly boosts the performance on abbr-100 for both LLM models. However, we also observe that neither Granite-13b-chat-v2.1 nor Llama2-13b-chat achieved a 1.0 recall score, even though we verified that the abbreviation matching the query was included in the augmented prompt prior to the query text. Currently, we conclude that some intrinsic limitations in LLMs, exacerbated by complex RAG-based contexts, prevent them from successfully recalling all abbreviations. We also present results for cmds-100 and q2a-100 after incorporating the ADH component, demonstrating that augmenting the prompt with additional abbreviation knowledge does not adversely affect the performance on these two evaluation datasets.

\section{Discussion}

In this paper, we demonstrate that using RAG leads to significantly superior outcomes compared to not using RAG. We also show that hybrid RAG exhibits noticeable enhancements over both sparse-only and dense-only RAG. One of our future directions involves fine-tuning more sophisticated sparse \cite{formal2021splade, gao-etal-2021-coil} and dense \cite{shi-etal-2022-improving} retrieval models  to enhance RAG even further. We will also explore improving the LLM model further through extended fine-tuning on our design data. Additionally, we aim to leverage reinforcement learning from human feedback (RLHF) \cite{ouyang2022training} technique to align our chat agent more closely with human preferences, based on the feedback data we collected from Slack GUI (shown in Fig. \ref{fig:interface}) responses.

\section{Conclusion}
In this work we introduce Ask-EDA, a chat agent tailored to boost productivity among design engineers. Empowered by LLM, hybrid RAG, and abbreviation de-hallucination techniques, Ask-EDA delivers relevant and accurate responses. We assessed Ask-EDA's performance across three distinct datasets, demonstrating its effectiveness across diverse domains. Finally, we integrated the Slack API to create a user-friendly natural language interface for seamless interactions with our chat agent.

\section{Acknowledgement}
We would like to express our sincere gratitude to Ehsan Degan, Vandana Mukherjee, and Leon Stok for their invaluable support and guidance as management throughout the preparation of this paper.

\bibliographystyle{IEEEtran}
\bibliography{IEEEabrv,main}

\appendices
\setcounter{figure}{0}
\renewcommand\thefigure{\Alph{section}.\arabic{figure}}

\section{Chat Interface}\label{app:Chat Interface}
\begin{figure}[!t]
    \centering
    \includegraphics[width=0.34\textwidth]{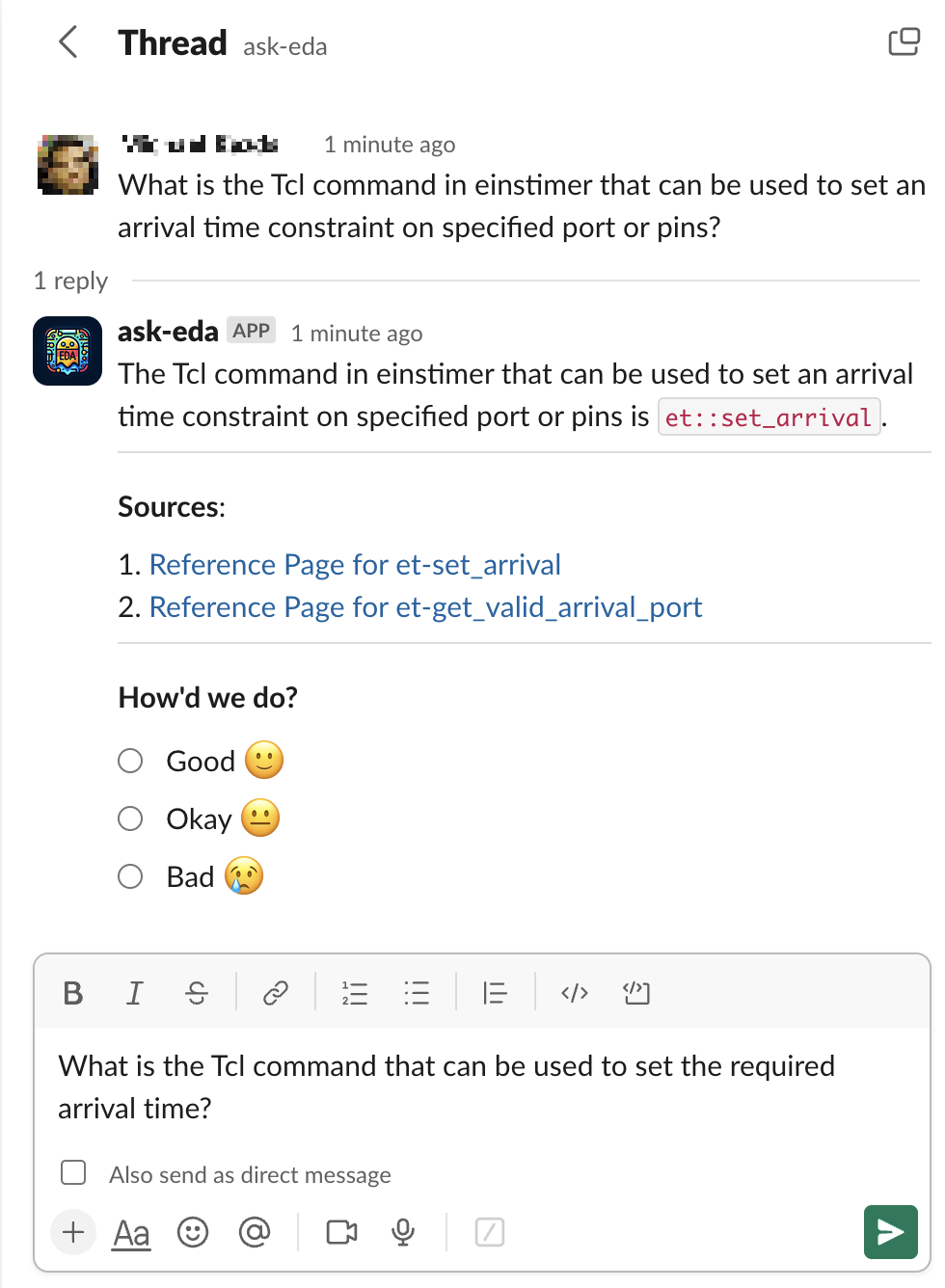}
    \caption{Example screenshot of Slack interface.}
    \label{fig:interface}
\end{figure}

Fig. \ref{fig:interface} shows an example screenshot of our Slack interface. The interface enables users to review the sources provided by RAG and give feedback on the quality of Ask-EDA’s response.

\end{document}